# Ontology based system to guide internship assignment process

Abir M'baya[1], Jannik Laval[1], Nejib Moalla[1], Yacine Ouzrout[1], Abdelaziz Bouras[2]

[1]Université de Lyon, Université Lumière Lyon 2, DISP Lab EA4570
Lyon, France

[2]CSE Department, Qatar University, Doha, Qatar

[1]FirstName.LastName@univ-lyon2.fr, [2]FirstName.LastName@qu.edu.qa

*Abstract*—Internship assignment is a complicated process for universities since it is necessary to take into account a multiplicity of variables to establish a compromise between companies' requirements and student competencies acquired during the university training. These variables build up a complex relations map that requires the formulation of an exhaustive and rigorous conceptual scheme. In this research a domain ontological model is presented as support to the student's decision making for opportunities of University studies level of the University Lumiere Lyon 2 (ULL) education system. The ontology is designed and created using methodological approach offering the possibility of improving the progressive creation, capture and knowledge articulation. In this paper, we draw a balance taking the demands of the companies across the capabilities of the students. This will be done through the establishment of an ontological model of an educational learners' profile and the internship postings which are written in a free text and using uncontrolled vocabulary. Furthermore, we outline the process of semantic matching which improves the quality of query results.

## I. INTRODUCTION

The collaboration between universities and companies has a great impact in educational and training systems for preparing young people to the job market. Cooperative training programs are looked as a vital resource for students to gain the required skills for employment carriers.

In this field, many universities provide an alternative education or "co-operative" courses or a work-link training in which the students attend for an academic period and then they can put immediately their knowledge acquired into practice. ULL follows this original mode of training such it helps students to develop their potentials through their practice in companies. Consequently, a recommender system must be deployed in ULL in order to establish a compromise between the concrete needs of companies in term of human resources and the competencies of students, i.e. this system helps companies in recruiting top talents and students in selecting the appropriate internship which corresponds to their skills and interests and building more easily their professional project.

An internship assignment is a hard process because the internship posting must corresponds to the qualifications of the student and its skills acquired during its academic background. Therefore, the top talents are selected by many companies, others don't find the appropriate internship and the passable qualifications may not be selected by any company. To resolve these problems, we would like to enrich a basic knowledge of missions or internship postings and learner candidate's profiles that succeed their internships. This is would be achieved in order to set up a recommender system that helps universities to make decisions about the internship assignment for students. To do so, we firstly compute the similarity between current missions with the previously mission's instances in the knowledge base. The profile of student that corresponds to the previously mission which has the highest similarity with the current internship posting is returned. Then, we calculate the similarity between the returned profiles with current learner student profiles. Finally, the profiles having the highest similarity are recommended to the current mission. In this way, missions and candidates' information must be stored in a hierarchical structure. We therefore propose in this paper a semantic internship assignment process based on ontology which is associated specification with semantic annotation. A semantic recommender system brings semantic to the ordinary recommender system with formalized knowledge and data that helps universities to deal with.

This paper is structured as follows. In section II, we review related works. In section III domain ontology of three main parts: Profile of the student, training education and internship posting are defined and constructed. In section IV, we outline the semantic matching approach. Finally, section V concludes the paper and our future works.

## II. RELATED WORKS

Domain ontology emerged in mainstream in many applications. There are some important previous and related works developed about the user profile and context ontology-based approach. [1], [2], [3], and [4] are related with Decision Support for the same educational domain. Recently it has increased the importance of applying ontology as a key part of efficient filtering in recommendation systems [5], [6].

From a recruitment process's viewpoint, Job adverts can be used as a source of information about the details of the employer demand including the qualification and skill requirements in the job market [12]. Although employment is still mostly a subject of highly level analysis, a lot of previous works presented by various authors were documented with the main purpose to extract information about skill demands through implementing machine learning algorithms and applying them to publicly available data. [13] uses text mining to extract and map skills listed in job postings to defined occupations. The work does neither implement an automatic pipeline nor an extensive evaluation to prove the proposed approach. [14] deals with skills required for Business Intelligence and Big Data jobs by finding the patterns to discover similarities and differences using the Latent Semantic Analysis (LSA) and Singular Value Decomposition (SVD). In [15], a cluster analysis in two phases was demonstrated to analyze job adverts based on current skill sets by applying cluster analysis with hierarchical agglomerative clustering and k-means. In [14], a skill system is established for generating occupational competencies with an automated approach using Named Entity Recognition (NER) and Named Entity Normalization (NEN) for raw texts.

Other several efforts focus on matching the right job with the right candidate by setting up a recommender system. The recommender systems usually compare the collected data with similar data collected from others and calculate a list of recommended items for the user. To do so, recommender techniques such as content-based filtering, Rule-based filtering, Collaboration filtering and Hybrid filtering can be applied [2].

Related studies believe that interactions are important for recommendation [5] as they have a great impact on the candidate's job choice and employer's hiring decision. Some interaction-based recommendation systems, such as CASPER [7], make use of collaborative filtering to recommend jobs to users based on what similar users have previously liked. Hybrid systems are also applied to match people skills and jobs description offer by taking into account both the preferences of the recruiters and the interests of the candidates.

However, there has been relatively little research exploring semantic balanced matching systems between the concrete needs of the companies and employee's competencies. To the best of our knowledge, no previous published work has applied a recommender system integrating semantic information related to the student competencies and interests, training courses and job postings in order to draw a balance taking the requirements of companies across the skills of the students.

In the context of an alternative education, we want to provide a sematic information system based on ontological models that allows us to better capture, analyze and use relevant semantic information for the exploitation and the simultaneous assignment of the internships to the students. Besides, our system will exploit the relationship between the demands of the missions and the student's profile who have followed a number of training courses and have some experience in finer detail. Consequently, the proposed system helps universities to satisfy both companies and students.

## III. CONSTRUCTION OF ONTOLOGY MODELS

In the first step in realizing our recommender system, we have created our different ontology models: mission ontology, student profile ontology and training courses ontology by integrating some widespread standards and classifications.

### A. Ontology concept

Ontology is an explicit formal specification of a shared conceptualization of a given domain of interest [9]. It defined a complex relations map that requires to be formulated in an exhaustive and rigorous conceptual scheme to constitute a knowledge base by capturing a shared meaningful of terms. It leans on this perspective to show understanding information (semantic). According to [10], this form of knowledge is considered as an intermediate representation of a conceptualization that is more formal and structured than the natural language, but less formal than a formal language, which allows establishing a common language which can be understood and capture the accumulated knowledge. OWL is the widely accepted approach to standardize a language for ontologies. OWL ontology consists of Individuals, Properties and Classes. Individuals (also known as instances) represent objects within a given domain. Properties state relationships between individuals or between individuals and data values. OWL class defines a group of individuals that share certain properties [8]. In this way, the data are referenced by metadata, under a standard normalized scheme which represent an abstract model of the real word. This mode of representation allows the interchange data following these standard schemes and, even, they can be modified and reused.

In our study we are interested in application ontology and we precise its mean throughout this paper as an ontology that captures more and more semantics from input models. The ontological models provide us a support in our decision system for analyzing and assessing information by taking into account the evolution of the concrete needs of the company and the student's skills.

### B. The mission ontology model

Nowadays, internship descriptions are written in form of free text using uncontrolled vocabulary. In contrast, semantic annotation of internship postings using concepts from a controlled vocabulary, based on Semantic Web technologies leads to have a standardise structure of mission's descriptions and consequently a better matching of student's skills and internship postings. The mission ontology is shown in figure 1.

The mission ontology is composed of many classes described as follows:
- *Location*: relate to the address of the mission
- *Competencies*: presents the competences required by the company. It is composed of two subclasses which are *Action* and *Domain-Action*.
  a) *Action:* presents the keywords (verbs) that describe the actions required to do at the internship.
  b) *Domain-Action:* presents the domain of the actions.
  
  The compositions of the two instances of the action and the domain-action help us to specify the student's diploma required by the company.
- *Experience*: define the required student's experiences.
- *Activity-Area*: specify the area of activities required in the internship
- *Project*: represents the project of the mission
- *Tasks*: relate to the tasks that must be done in the project of the mission

- *Time*: defines the duration of the mission, the start and end date of the tasks.
- *History*: presents the years of partnership of the company with the university, the total number of missions done in the company and the number of missions done with difficulties by the students.
- *Company*: define the name of company, the company importance and the number of employees.
- *Assignment*: represent the assignment process in which we will position one or more students in a mission.
- *Argument*: it's the proof that allows the university to justify the assignment. They are detailed in section IV.

*C. The student profile ontology*

The student profile is created by taking all information from the student. It consists of the classes' definitions based on shared attributes, emphasizing on information needs, access conditions, experience, competencies and knowledge. In this section, we extract all the information related to the student of ULL and then we construct our student profile ontology model. The student information is broadly divided into many concepts as shown in figure 2.

- *Student-Role*: this class presents the role that can have a student and it is composed of the sub-class Delegates.
- *Student-Status*: relate to the training that follows the student in ULL university. This class consists of three sub-classes VAE, Initial-training, Continuous-training.
- *Academic-information*: details the academic information such as the degree of the student and the actual academic year.
- *Administrative-information*: contains the personal details of the student such as its first and last name, number phone, address, email, nationality and its age.
- *Evaluation-record*: contains the information about the level-professional of the students i.e. the notes of its oral presentation, its quality of work and its behavior obtained during its internship in the company. Its evaluation-record is evaluated by the member of both company and university.
- *Candidate-record*: contains the information about the academic level of the student by evaluating him in some items such as quality of its experiences, its knowledge of the field of project management and monitoring process, overall rating of its curricula vitae, etc.
- *Interests*: they express the preferences of the students in terms of mission, location, salary and company.

*D. The University ontology*

University ontology is shown in figure 3. It is important in the design of our solution because it allows us to make a link between students and companies and to define some rules and constraints. It is precisely within the university that students upgrade their skills in order to reach the highest level that would enable them to succeed in the professional world. Because of its density, we will present in this article only some concepts of this ontology that are necessary for the understanding and the conceptualization of our system.

First of all, *University* is composed of *departments* that organize trainings. Each *training* includes *teaching units* that are mainly characterized of objectives, target competencies and keywords associated with them. The teaching unit is itself composed of *modules*, enabling to inform about the importance of the courses that will be taught to the student into a module, through the number of hours, the ECTS credits and the coefficient. The *course* contains a finer level of competencies representation through keywords and the number of hours. The course is taught by a *teacher* that is part of the *Teaching Staff*.

Then, we can see that it's the department that creates a *partnership* in order to be in relation with a *company* that will share some internship offers with the university. Theses missions will be offered to the students after the *assignment* process. This one will respect the *constraints* created by the University and take into account several criteria such as student level at the courses that best fits the mission requirements. The course level could be represented by the *mark* obtained by the student. The constraints are the university requirements that have to be taken into account during the selection process, like the minimum and maximum number of students that have to be presented to the company in response to the mission offer.

IV. SEMANTIC MATCHING

As we described above, the internship assignment is a very complicate process. The main raison is that the companies want to have the best qualifications which correspond to their concrete needs. So, the top talents of students are selected by several companies. In contrast, some of students don't find the appropriate internship that interests them and some others don't find any internship and consequently they can't validate their degrees.

To resolve this problem, a recommended system must be deployed to assign an internship to every student. As we say above, this system consists of balancing of company's needs described in missions and student's capabilities.

Having an internship offer, a student profile and a training courses described using controlled vocabulary from our ontology models allows us to perform semantic matching. These ontological models provide a support for our complex problem-solving and facilitate knowledge modelling and reuse. Besides, there are capable to analyze and assess our decision support system by taking into account the evolution of the concrete needs of the company and the student's skills.

Figure 4 describes the selection-decision made by university about the assignment of the missions offer to the students. This decision that respect some constraints defined by university must match efficiently between the company's requirements and student's skills. The returned results are accompanied by arguments in order to convince companies to select the proposed student by the university.

The arguments are defined in the following way:
- A1_Same_Old_Student_Profil_For_Similar_Mission: The proposed student has a profile similar to a former student who has successfully completed his mission.
- A2_Sufficient_Level_For_Competencies_Of_Mission: The student has sufficient level in the courses (competencies) required in the mission.
- A3_Standard_Student_For_A_Standard_Mission: The mission is standard because it is similar to the standard mission profile. It is known in the experience of supervisors and it does not present any risks.
- A4_Student_With_High_General_Level: The student has a high level.
- A5_Student_Motivated_matching_with_his_interests: The student is motivated because there is a good matching between his interests and the demands of company.
- A6_Not_Perfect_Matching_But_Student_With_Great_Autonomy: There is no perfect matching of skills but this student demonstrates a considerable autonomy, so he can learn quickly.

To set up our recommender system, we will first start by structuring the missions' offers that is written in form of free text into formal annotation. We then extract the keywords of the internship posting and integrate them into missions' concepts in order to create missions' instances. These retrieval data are constructed and maintained by a domain expert. Moreover, by applying a clustering algorithm, we regroup the most similar missions in term of their structure and semantic annotation into a same cluster. The similarity function between missions' instances is computed by using the cosine similarity of the Vector Space Model (VSM) which is constructed by the identified missions' concepts. In our approach, the concepts that identify VSM are Action, Domain of the action and activity-area.

Therefore, we will identify the profile of the previous student who has succeeded his internship from each mission in order to find the types of profiles which correspond to each cluster of missions. To do so, we will select the student who has the most similar profiles to the previous student. As a returned result, we will have a cluster of students that corresponds to each cluster of missions. This result constitutes our knowledge base that helps us to extract relevant semantic rules and learn the type of profiles that corresponds to the internship posting.

Moreover, the exploitation of our knowledge base allows us to determine the appropriate internship to the new candidate.

We therefore classify the current internship posting/current profile of the student into their appropriate clusters by computing their degree of semantic similarity with the previous ones that constitute the knowledge base. Consequently, the complex decision-making process returns in a narrower problem. In fact, for a given internship posting a ranked list of the best matching candidates of the associated group can delivered as a result, and vice versa. The final similarity between candidate's profile and internship offer calculate the degree similarity between the keywords describing the missions and the acquired skills of the student.

This result is considered as a first assignment of the students to the appropriate mission. To optimize it by drawing a balanced distribution taking the demands of the companies across the students' capacities, we apply a multi-criteria optimization algorithm such as genetic algorithm.

## V. IMPLEMENTATION OVERVIEW

We have chosen Protégé tool based on the ontological language in order to implement our ontological models, due to its extensibility, quick prototyping and application development. Protégé ontologies are easily exported into different formats including RDF Schema, Web Ontology Language (OWL) which is defined as a language of markup to publish and to share data using ontologies in the WWW and Top Braid Composer.

To establish a balanced distribution between the companies' needs and the students' skills, we will develop a java implementation in order to exploit the knowledge base created by instantiating our ontological models. This program allows the user to query the news and view the knowledge base. It uses the framework Jena which provides integrated implementations of the W3C Semantic Web Recommendations, centered on the RDF graph for manipulating and reasoning with the OWL ontologies. For querying, it employs SPARQL and tSPARQL [11], which adds time functionalities to the queries. Then, we will use the Weka framework to implement our clustering and optimization algorithms. Finally, the results will be returned in a web interface.

## VI. CONCLUSION

In this paper, we proposed a recommender system in the assignment internship process. We described the ontological models, the student profile, training course and internship posting, used within our approach which provides us means for semantic annotation. Using controlled vocabularies, in contrast to free text descriptions, results in a better machine process ability, data interoperability and integration. Moreover, having internship offer and student profiles semantically annotated, enables us to perform semantic matching which significantly improves query results and delivers a ranked list of best matching candidates for a given internship position.

## VII. ACKNOWLEDGMENT

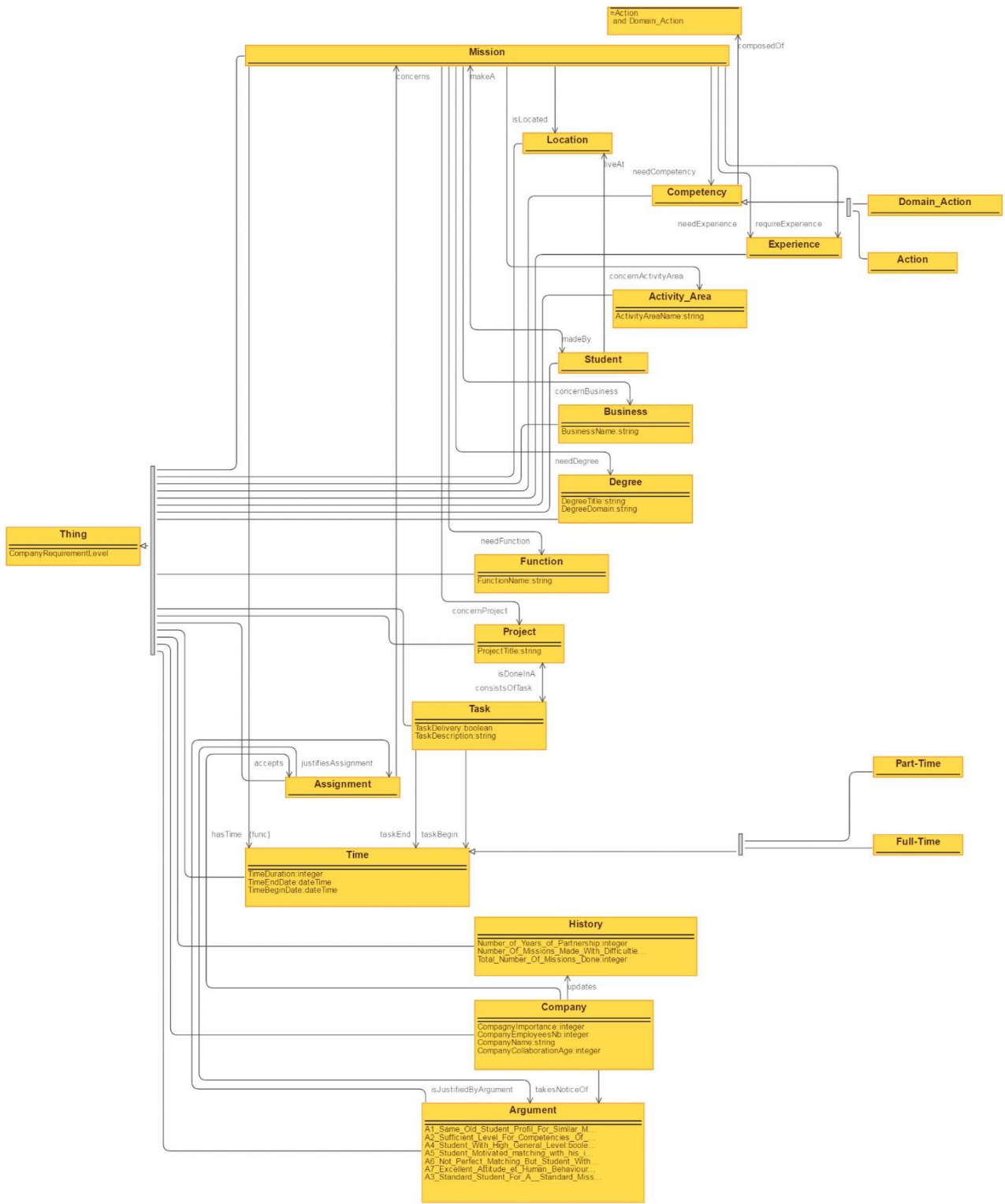

Figure 1. The mission ontology model

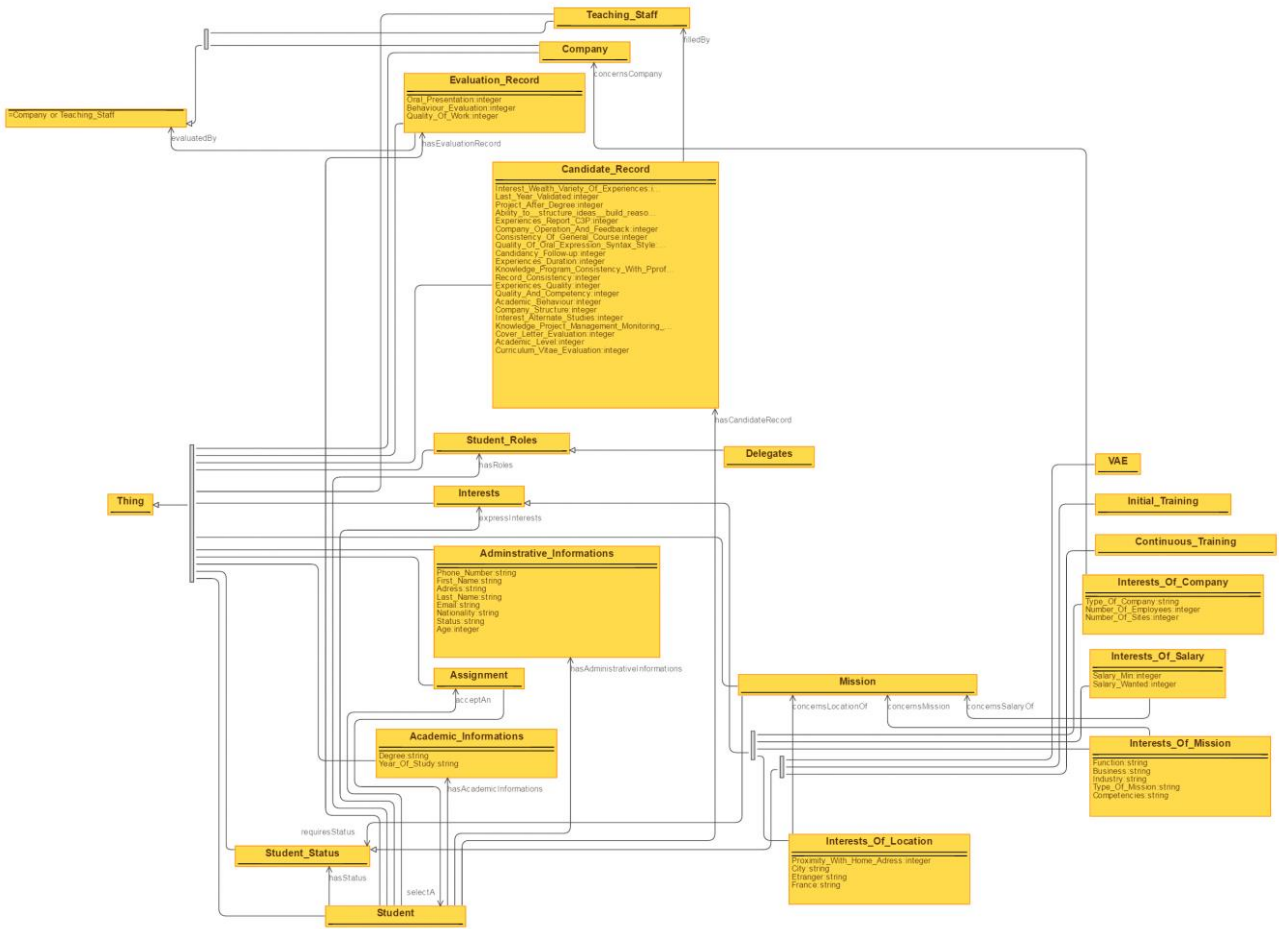

Figure2. The student profile ontology model

Figure3. The University ontology model

Figure 4: The selection-decision of assignment process